\documentclass[10pt,twocolumn,letterpaper]{article}

\usepackage{cvpr}
\usepackage{times}
\usepackage{amsmath}

\usepackage{graphicx}
\usepackage{caption}
\usepackage{amsmath,amssymb,enumerate}

\usepackage{amsthm}
\usepackage{mathtools}
\usepackage{breqn}
\usepackage{algorithm, algorithmicx, algpseudocode}

\usepackage{blindtext}
\usepackage{gensymb}
\usepackage{xparse}
\usepackage{lipsum}
\usepackage{mathrsfs}
\usepackage[mathscr]{euscript}
\usepackage{times}
\usepackage{cite} 
\usepackage{multicol}
\usepackage[caption=false,font=footnotesize]{subfig}
\usepackage{amsfonts}
\usepackage[utf8]{inputenc}
\usepackage[T1]{fontenc}
\usepackage{textcomp}
\usepackage{amsfonts}
\usepackage{soul}
\usepackage{multirow}
\usepackage{booktabs}
\usepackage[usenames,dvipsnames,svgnames,table, xcdraw]{xcolor}

\newtheorem{problem}{Problem}

\DeclareMathOperator*{\argmax}{arg\,max}
\DeclareMathOperator*{\argmin}{arg\,min}
\DeclareMathOperator{\tr}{Tr}

\newcommand{\m}{\mathop{\mathrm{m}}}

\newcommand{\Hcal}{\mathcal{H}}
\newcommand{\Ical}{\mathcal{I}}

\newcommand{\transpose}{\mathsf{T}}

\DeclareDocumentCommand{\vectorToSkew}{ O{} }{\left(#1\right)_\times}

\usepackage[breaklinks=true,bookmarks=false]{hyperref}

\cvprfinalcopy 

\def\cvprPaperID{****} 

\begin{document}

\title{A Keyframe-based Continuous Visual SLAM for RGB-D Cameras via Nonparametric Joint Geometric and Appearance Representation}

\author{Xi Lin, Dingyi Sun, Tzu-Yuan Lin, Ryan M. Eustice, and Maani Ghaffari\\
University of Michigan, Ann Arbor, MI 48109\\
{\tt\small \{bexilin,dysun,tzuyuan,eustice,maanigj\}@umich.edu}
}

\maketitle

\begin{abstract}
   This paper reports on a robust RGB-D SLAM system that performs well in scarcely textured and structured environments. We present a novel keyframe-based continuous visual odometry that builds on the recently developed continuous sensor registration framework. A joint geometric and appearance representation is the result of transforming the RGB-D images into functions that live in a Reproducing Kernel Hilbert Space (RKHS). We solve both registration and keyframe selection problems via the inner product structure available in the RKHS. We also extend the proposed keyframe-based odometry method to a SLAM system using indirect ORB loop-closure constraints. The experimental evaluations using publicly available RGB-D benchmarks show that the developed keyframe selection technique using continuous visual odometry outperforms its robust dense (and direct) visual odometry equivalent. In addition, the developed SLAM system has better generalization across different training and validation sequences; it is robust to the lack of texture and structure in the scene; and shows comparable performance with the state-of-the-art SLAM systems.  
\end{abstract}

\section{Introduction}

Visual SLAM has been a focused research topic and widely applied to areas like 3D reconstruction~\cite{Zollhofer2018recon,Newcombe2011KF}, augmented reality~\cite{klein2007AR,Newcombe2010denserec}, and mobile robotics~\cite{huang2017visual,Stefan2015kfno}. The feature-based SLAM has long been regarded as the mainstream method~\cite{Schonberger2016sfm,Endres2012eval}.
Key strengths for the feature-based SLAMs are their computational efficiency and reliable place recognition performance. However, using a sparse set of features comes at the cost of discarding most of the image information, resulting in tracking lost and lack of robustness.
\begin{figure}
    \centering
    \includegraphics[width=0.99\columnwidth]{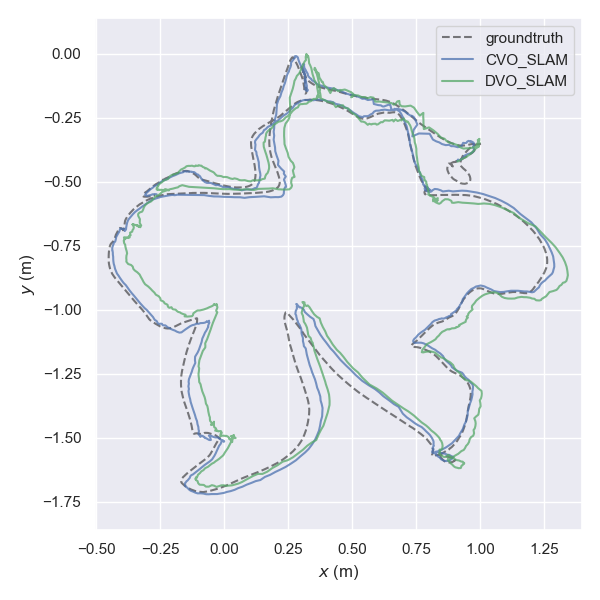}
    \caption{An example of the proposed keyframe-based continuous visual SLAM compared with DVO SLAM~\cite{kerl2013dense} on TUM \texttt{fr1/plant} sequence~\cite{TUM_dataset}.}
    \label{fig:first_fig}
\end{figure}

Direct methods have been recently successful in odometry and SLAM systems~\cite{audras2011real,Steinbrucker2011DVO,kerl2013robust,wang2017stereo,engel2018direct,engel2014lsd}. The fundamentally novel formulation of the sensor registration problem in~\cite{MGhaffari-RSS-19} is continuous and models the action of an arbitrary Lie group on any smooth manifold. In this approach, unlike direct energy formulation there is no need for constructing an image pyramid and solving the same registration problem at different resolutions.  

In this paper, building on the Continuous Visual Odometry (CVO) in~\cite{MGhaffari-RSS-19}, we propose a robust RGB-D SLAM system that performs well in scarcely textured and structured environments.
In particular, this work has the following contributions:
\begin{enumerate}
    \item We develop an intrinsic keyframe selection technique using the continuous visual odometry that directly operates on 3D colored point clouds. We show that the proposed technique is effective and can improve tracking performance. The keyframe-based continuous visual odometry is extended to a SLAM system. We use loop-closure constraints that combine indirect loop-closure detection via ORB features and direct transformation estimation with continuous visual odometry. 
    
    \item We present extensive evaluations on publicly available benchmarks and show that the developed keyframe selection technique using continuous visual odometry outperforms its equivalent robust dense (and direct) visual odometry. Furthermore, the SLAM system has better generalization across different training and validation sequences; it is robust to the lack of texture and structure in the scene; and shows comparable performance with the state-of-the-art SLAM systems. 
    
    \item An open-source implementation of this work is available for download at: {\small \href{https://bitbucket.org/perl-sw/cvo-slam/src/master/}{\url{https://bitbucket.org/perl-sw/cvo-slam/src/master/}}}
\end{enumerate}

The remaining of this paper is organized as follows. A review of related work is given next. A background on continuous sensor registration is given in Section~\ref{sec:preliminaries}. Section~\ref{sec:kfcvo} presents the keyframe-based continuous visual odometry. Sections~\ref{sec:loop_closure} and \ref{sec:posegraph} discuss the extension of the keyframe-based odometry to a SLAM system via loop-closure constraints generation and robust pose graph optimization. The experimental results and discussion are presented in Section~\ref{sec:exp}. Finally, Section~\ref{sec:conclusion} concludes the paper and shares future work ideas.

\section{Related work}
 
Henry \textit{et al.}~\cite{Henry2010rgbicp} proposed an RGB-D Iterative Close Point (ICP) method. The RGB-D-ICP initializes the sensor motion with SIFT feature matching and utilizes the ICP algorithm to estimate the pose. The SIFT features are also used for loop-closures detection. A pose graph optimization is applied to maintain global consistency. In KinetFusion, Newcombe \textit{et al.}~\cite{Newcombe2011KF} proposed 3D scene reconstruction using depth cameras and the truncated signed distance function. The camera pose is tracked using a coarse-to-fine ICP by aligning each frame to the global model. The align and fuse strategy associates the current frame with measurements from the past, thereby establishes the loop-closure and reduces the drift compared with the frame to frame tracking. Whelan \textit{et al.} extended KinetFusion to an RGB-D system~\cite{whelan2013robust}. Elastic Fusion also used the same frame-to-model tracking approach~\cite{Whelan2016elastic}. Surface loop-closure is applied frequently to maintain the consistency of the map. 

In~\cite{audras2011real,Steinbrucker2011DVO,kerl2013robust}, direct image alignment is achieved by minimizing the energy function that corresponds to the photometric error. Kerl \textit{et al.}~\cite{kerl2013robust,kerl2013dense,Kerl2013DVOrepo} combined this method with a keyframe-based pose graph SLAM, where the Gaussian entropy ratio was proposed as a metric of similarity between frames. The keyframe selection and the loop-closure validation are based on the hard thresholding of such a ratio. As for loop-closure detection, they choose the metrical nearest-neighbor search. While the reported performance was good, this strategy is only functional within a small indoor environment with sufficiently accurate odometry estimation, which can make the application of the proposed system restricted.

The recent BAD SLAM presents a fast direct Bundle Adjustment (BA) formulation~\cite{schps2019bad}. Starting from the keyframe-based direct RGB-D tracking, the BAD SLAM creates dense surfels for each keyframe. The surfels are associated with descriptors and tracked carefully with BA. The BAD SLAM can achieve high-accuracy trajectory estimation on sequences recorded with a well-calibrated global shutter camera. However, as the number of keyframes increases, the number of tracked surfels grows linearly, and the system runtime violates the real-time requirement within a minute. Moreover, even with the ideal camera setup, this method is volatile to motion blur. As soon as the image gets blurry, the method diverges, and recovery rarely can happen.

\section{Background on continuous sensor \\ registration}
\label{sec:preliminaries}

Consider two (finite) collections of points, $X=\{x_i\}$, $Z=\{z_j\}\subset \mathbb{R}^3$. We want to determine which element \mbox{$h\in \mathrm{SE}(3)$}, where \mbox{$R \in \mathrm{SO}(3)$} and \mbox{$T \in \mathbb{R}^3$}, aligns the two point clouds $X$ and $hZ = \{hz_j\}$ the ``best.'' To assist with this, we will assume that each point contains information described by a point in an inner product space, $(\mathcal{I},\langle\cdot,\cdot\rangle_{\mathcal{I}})$. To this end, we will introduce two labeling functions, $\ell_X:X\to\Ical$ and $\ell_Z:Z\to\Ical$.

In order to measure their alignment, we will be turning the clouds, $X$ and $Z$, into functions $f_X,f_Z:\mathbb{R}^3\to\Ical$ that live in some reproducing kernel Hilbert space, $(\Hcal,\langle\cdot,\cdot\rangle_{\mathcal{H}})$. The action, $\mathrm{SE}(3) \curvearrowright \mathbb{R}^3$ induces an action $\mathrm{SE}(3) \curvearrowright C^\infty(\mathbb{R}^3)$ by $h.f(x) := f(h^{-1}x)$. Inspired by this observation, we will set $h.f_Z := f_{h^{-1}Z}$.

\begin{problem}\label{prob:problem}
	The problem of aligning the point clouds can now be rephrased as maximizing the scalar products of $f_X$ and $h.f_Z$, i.e., we want to solve
	\begin{equation}\label{eq:max}
		\argmax_{h\in \mathrm{SE}(3)} \, F(h),\quad F(h):= \langle f_X, h.f_Z\rangle_{\mathcal{H}}.
	\end{equation}
\end{problem}

We follow the same steps in~\cite{MGhaffari-RSS-19} with an additional step in which we use the kernel trick to kernelize the information inner product. For the kernel of our RKHS, $\Hcal$, we first choose the squared exponential kernel $k:\mathbb{R}^3\times\mathbb{R}^3\to\mathbb{R}$:
\begin{equation}\label{eq:k}
k(x,z) = \sigma^2\exp\left(\frac{-\lVert x-z\rVert_3^2}{2\ell^2}\right),
\end{equation}
for some fixed real parameters (hyperparameters) $\sigma$ and $\ell$, and $\lVert\cdot\rVert_3$ is the standard Euclidean norm on $\mathbb{R}^3$. This allows us to turn the point clouds to functions via
\begin{align}
	\nonumber f_X(\cdot) &:= \sum_{x_i\in X} \, \ell_X(x_i) k(\cdot,x_i), \\
	f_Z(\cdot) &:= \sum_{z_j\in Z} \, \ell_Z(z_j) k(\cdot,z_j).
\end{align}
We can now define the inner product of $f_X$ and $f_Z$ by
\begin{equation}\label{eq:scalar}
\langle f_X,f_Z\rangle_{\Hcal} := \sum_{\substack{x_i\in X, z_j\in Z}} \, \langle \ell_X(x_i),\ell_Z(z_j)\rangle_{\mathcal{I}} \cdot k(x_i,z_j).
\end{equation}

We use the well-known kernel trick in machine learning~\cite{bishop2006pattern,rasmussen2006gaussian,murphy2012machine} to substitute the inner products in~\eqref{eq:scalar} with the appearance (color) kernel. The kernel trick can be applied to carry out computations implicitly in the high dimensional space, which leads to computational savings when the dimensionality of the feature space is large compared to the number of data points~\cite{rasmussen2006gaussian}. After applying the kernel trick to~\eqref{eq:scalar}, we get
\begin{align}
\label{eq:newscalar}
    \nonumber \langle f_X,f_Z\rangle_{\Hcal} =& \sum_{\substack{x_i\in X, z_j\in Z}} \,  k_c(\ell_X(x_i),\ell_Z(z_j)) \cdot k(x_i,z_j) \\
    :=& \sum_{\substack{x_i\in X, z_j\in Z}} \,  c_{ij} \cdot k(x_i,z_j),
\end{align}
where we choose $k_c$ to be also the squared exponential kernel with fixed real hyperparameters $\sigma_c$ and $\ell_c$ that are set independently.

\section{Keyframe-based continuous visual odometry via inner product ratios}
\label{sec:kfcvo}

\begin{figure*}[t]
\begin{center}
\includegraphics[trim=0cm 0.55cm 0cm 0cm,clip,width=1.99\columnwidth]{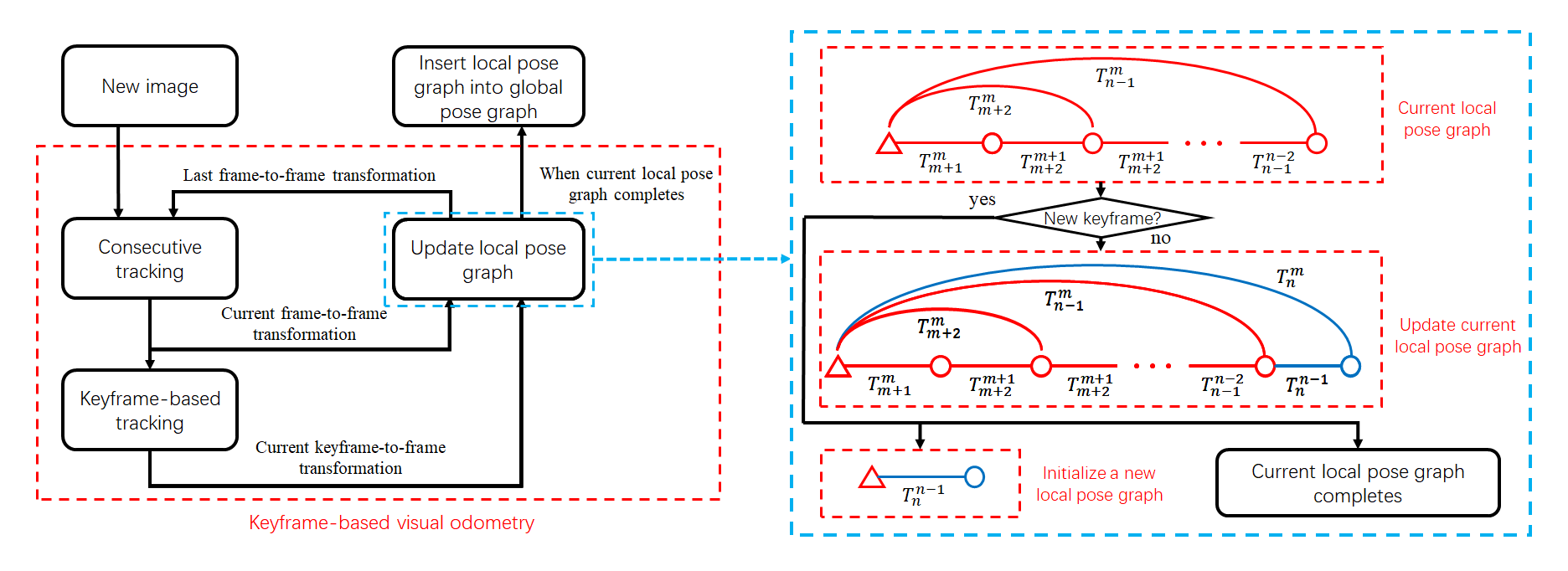}
\end{center}
   \caption{Pipeline for keyframe-based continuous visual odometry. The left part shows the flow of keyframe-based odometry, which takes a new image as input and outputs a complete local pose graph to the global pose graph. The right part illustrates how the local pose graphs are updated. If the current frame is not a new keyframe, then a new node (blue circle) is created for the current frame and the transformations at the current frame are set to be new edges (blue lines). Otherwise, the current local pose graph will be added to the global pose graph and a new local pose graph will be initialized.}
\label{fig:Keyframe-based visual odometry pipeline}
\end{figure*}

When performing consecutive frame odometry, i.e., \emph{frame-to-frame} odometry, the total trajectory drift is accumulated at every frame. To reduce the total drift, we adopt the idea of keyframe-based tracking~\cite{vacchetti2003fusing,klein2007parallel,klein2008improving,kerl2013dense}. Instead of estimating the global pose at the current frame with respect to that of the last frame, the current pose is computed using the current keyframe pose and \emph{keyframe-to-frame} odometry.

The overall pipeline of the developed keyframe-based odometry is shown in Figure~\ref{fig:Keyframe-based visual odometry pipeline}. The right part of the figure shows a detailed illustration of the steps  "Update local pose graph according to tracking quality," where the notations are defined in the following paragraphs. 

\subsection{Keyframe-based transformation estimation}

When receiving a new image $n$, we first track it with respect to the last frame $n-1$ to get the consecutive transformation $T_n^{n-1}$ by solving an instance of Problem~\ref{prob:problem}:

\begin{equation}\label{eq:consecutive transform}
    T_{n-1}^n= \argmax_{h\in \mathrm{SE}(3)} \,  \langle f_{X_{n-1}}, h.f_{X_n}\rangle_{\mathcal{H}} ,\ h_0 = T_{n-2}^{n-1} ,
\end{equation}
where we denote $T_{n-1}^n = (T_n^{n-1})^{-1}$. In~\eqref{eq:consecutive transform}, $f_{X_{n-1}}$ and $f_{x_n}$ are point functions at the last frame $n-1$ and current frame $n$, and $h_0$ is the initial pose for the solver, which is set to the previous pose $T_{n-1}^{n-2}$. This assumption can also be interpreted as a constant velocity motion model for consecutive tracking initialization.

To get the transformation from the current keyframe $m$ to the current frame $n$, $T_n^m$, we similarly solve another instance of Problem~\ref{prob:problem}:

\begin{equation}\label{eq:keyframe-to-frame transform}
    T_m^n= \argmax_{h\in \mathrm{SE}(3)} \,  \langle f_{X_m}, h.f_{X_n}\rangle_{\mathcal{H}} ,\ h_0 = T_{n-1}^n T_m^{n-1}
\end{equation}

In~\eqref{eq:keyframe-to-frame transform}, we set the initial value $h_0$ using the current frame-to-keyframe pose $T_m^{n-1}$ and the consecutive pose $T_{n-1}^n$. Consequently, the estimated pose of the current frame in the world frame, $T_n^{\text{W}}$, can be obtained multiplying the current keyframe pose $T_m^{\text{W}}$ with the keyframe-to-frame pose $T_n^m$.

\begin{table*}[t]
    \centering
    \caption{The RMSE of Relative Pose Error (RPE) for \texttt{fr1} sequences. Because some frames are missing in the original video of \texttt{fr1/floor}, we skip the evaluation on that specific sequence. The trans. columns show the RMSE of the translational drift in $\mathrm{m}/\sec$ and the rot. columns show the RMSE of the rotational error in $\mathrm{deg}/\sec$. There's no corresponding validation datasets for \texttt{fr1/teddy}. The results show that KF-CVO out-performs other methods on both training and validation sets of \texttt{fr1}.\label{Keyframe-based odometry result}}
    \resizebox{\textwidth}{!}{
    \begin{tabular}{lcccccccc|cccccccc}
        \toprule
         & \multicolumn{8}{c}{Training} & \multicolumn{8}{c}{Validation} \\
        \midrule
          & \multicolumn{2}{c}{CVO~\cite{MGhaffari-RSS-19}} &  \multicolumn{2}{c}{KF-CVO} & \multicolumn{2}{c}{DVO~\cite{Kerl2013DVOrepo}} &
          \multicolumn{2}{c}{KF-DVO~\cite{kerl2013dense}} &\multicolumn{2}{c}{CVO~\cite{MGhaffari-RSS-19}} & \multicolumn{2}{c}{KF-CVO} & \multicolumn{2}{c}{DVO~\cite{Kerl2013DVOrepo}}&
          \multicolumn{2}{c}{KF-DVO~\cite{kerl2013dense}} \\
         Sequence & Trans. & Rot. & Trans. & Rot. & Trans. & Rot. & Trans. & Rot. & Trans. & Rot. & Trans. & Rot. & Trans. & Rot. & Trans. & Rot. \\
        \midrule
        fr1/desk    & 0.0486 & 2.4860 & \bf 0.0355 & \bf 2.1443 & 0.0387 & 2.3589 & 0.0497 & 4.5420 &
                      0.0401 & 2.0148 & 0.0403 & \bf 1.9305 & \bf 0.0371 & 2.0645 &
                     0.0374 & 2.2389 \\
        fr1/desk2   & 0.0535 & 3.0383 &  \bf 0.0452 & \bf 2.8263 & 0.0583 & 3.6529 & 0.0573 & 4.1695 &
                      0.0225 & 1.7691 & \bf 0.0213 & \bf 1.6477 & 0.0208 & 1.7416 & 0.0268 & 2.0467 \\ 
        fr1/room    & 0.0560 & 2.4566 & \bf 0.0465 & \bf 2.2822 & 0.0518 & 2.8686 & 0.0556 & 2.6157 &
                      0.0446 & 3.9183 & \bf 0.0379 & \bf 3.6709 & 0.2699 & 7.4144 & 0.0536 & 3.8048 \\
        fr1/360     & 0.0991 & \bf 3.0025 & \bf 0.0828 & 3.3401 & 0.1602 & 4.4407 & 0.0989 & 4.2443 &
                      0.1420 & 3.0746 & \bf 0.0830 & \bf 2.2434 & 0.2811 & 7.0876 & 0.0641 & 2.4565  \\
        fr1/teddy   & 0.0671 & 4.8089 & \bf 0.0534 & 5.6454 & 0.0948 & \bf 2.5495 & 0.0565 & 2.9151 &
                      n/a    & n/a    & n/a    & n/a    & n/a    & n/a & n/a    & n/a  \\
        fr1/xyz     & 0.0240 & \bf 1.1703  & \bf 0.0235 & 1.2133 & 0.0327 & 1.8751 & 0.0237 & 1.4481 &
                      0.0154 & 1.3872 & \bf 0.0121 & \bf 0.8666 & 0.0453 & 3.0061 & 0.0205 & 1.4294\\
        fr1/rpy     & 0.0457 & 3.3073 & 0.0356 & 3.1083 & \bf 0.0336 & \bf 2.6701 & 0.0418 & 4.0099 &
                      0.1138 & 3.6423 & 0.0985 & \bf 3.4638 & 0.3607 & 7.9991 & \bf 0.0813 & 5.4651 \\
        fr1/plant   & 0.0316 & 1.9973 & \bf 0.0212 & \bf 1.5208 & 0.0272 & 1.5523 & 0.0397 & 1.8611 &
                      0.0630 & 4.9185 & \bf 0.0559 & 4.7094 & 0.0660 & \bf 2.5865 & 0.0657 & 4.9080  \\
        \midrule
        Average    & 0.0532 & 2.7834 & \bf 0.0430 & 2.7601 & 0.0622 & \bf 2.7460 & 0.0533 & 3.2258 & 0.0631 & 2.9607 & \bf 0.0499 & \bf 2.6475 &  0.1544 & 4.5571 & \bf 0.0499 & 3.1928 \\
        \bottomrule
    \end{tabular}
    }
    \label{tab:fr1_tracking_results}
\end{table*}

\subsection{New keyframe selection}

For keyframe selection, we define a novel parameter that is intrinsic to CVO using the ratio of two inner products:

\begin{equation}\label{eq:inner_product_ratio}
\gamma := \frac{\langle f_{X_m},\ T_m^n.f_{X_n}\rangle_{\Hcal}}{\langle f_{X_m},\ T_m^{m+1}.f_{X_{m+1}}\rangle_{\Hcal}},
\end{equation}
where $f_{X_m}$ and $f_{X_n}$ have the same definition as those of~\eqref{eq:consecutive transform} and~\eqref{eq:keyframe-to-frame transform}. Since the frame next to the current keyframe $m+1$ is the closest frame to the current keyframe $m$, the images are highly similar and $T_{m+1}^m$ is expected to be more accurate than future frames. Therefore, point cloud $X_m$ and the aligned point cloud $T_{m+1}^m X_{m+1}$ are also expected to be more similar than future point clouds. As such, the inner product $\langle f_{X_m},\ T_m^{m+1}.f_{X_{m+1}}\rangle_{\Hcal}$ computed using~\eqref{eq:newscalar} serves as a reference value for evaluating the alignment quality between current keyframe $m$ and current frame $n$.

The inner product ratio $\gamma$ indicates the relative alignment quality of the current keyframe-to-frame transformation $T_n^m$. In addition, $\gamma$ is a relative quantity, which means it is independent of the scale variation in absolute inner product values. A consequence of this property is that we can set a fixed threshold $\gamma_{\text{thres}}$. 

In addition, we also want to keep consecutive keyframes to be close to each other spatially and to have similar view angles, which means that the difference between position and orientation of consecutive keyframe poses. i.e.\@ \mbox{$T_n^m = T_{\text{W}}^m T_n^{\text{W}}$}, should not be large. We use the absolute value of the misalignment angle, $\theta_{mn}$, computed through
\begin{equation}\label{eq:rotation angle} 
    \theta_{mn} = \arccos\left(\frac{\tr(R_{mn})-1}{2}\right)
\end{equation}
where $\tr(\cdot)$ denotes the trace of a matrix. For the translational difference we use the Euclidean norm of translation part of $T_n^m$, denoted as $\lVert t_{mn}\rVert_3$. We set a translation norm threshold $t_{\text{thres}}$ for $\lVert t_{mn}\rVert_3$ and a misalignment angle threshold $\theta_{\text{thres}}$ for $\theta_{mn}$. If any of them exceeds the threshold, the difference between two keyframes is considered too large; hence, a new keyframe is initialized.

In summary, we select a new keyframe that has good image alignment quality with respect to the last keyframe and acceptable pose difference with last keyframe. When $\gamma < \gamma_{\text{thres}}$ or $\lVert t_{mn}\rVert_3 > t_{\text{thres}}$ or $\theta_{mn} > \theta_{\text{thres}}$, we set the last frame $n-1$ to be a new keyframe $m+1$. 

\subsection{Local pose graph update}

Along with the tracking process, we also maintain and update a local pose graph, which is similar to that of DVO SLAM~\cite{kerl2013dense}. The pose graph structure and update process are shown in Figure~\ref{fig:Keyframe-based visual odometry pipeline}. Every time the keyframe-based tracking of the current frame is completed, a new node is created in the local pose graph. The pose $T_n^m$ is set to be the estimation of an edge between the keyframe and the current frame nodes. The edge between the last frame and the current frame nodes is set to $T_n^{n-1}$. When a new keyframe is created, the current local pose graph is complete and inserted into the global pose graph in the \emph{backend}. Then a new local pose graph is initialized.

\subsection{Evaluation}

Table~\ref{tab:fr1_tracking_results} shows the keyframe-based tracking results using TUM RGB-D Benchmark~\cite{TUM_dataset}. The proposed keyframe-based continuous visual odometry denoted KF-CVO, has, respectively, 19\% and 21\% improvement on translational drift over frame-to-frame CVO tracking~\cite{MGhaffari-RSS-19} in \texttt{fr1} training and validation sequences; on average, the rotational drifts are also better. Compared to the state-of-the-art dense visual odometry method DVO~\cite{Kerl2013DVOrepo} and keyframe-based DVO, which is used in DVO SLAM~\cite{kerl2013dense}, KF-CVO also has a better overall performance.

\section{Loop-closure detection and transformation estimation}\label{sec:loop_closure}

\begin{figure*}[t]
\centering
\includegraphics[width=2\columnwidth]{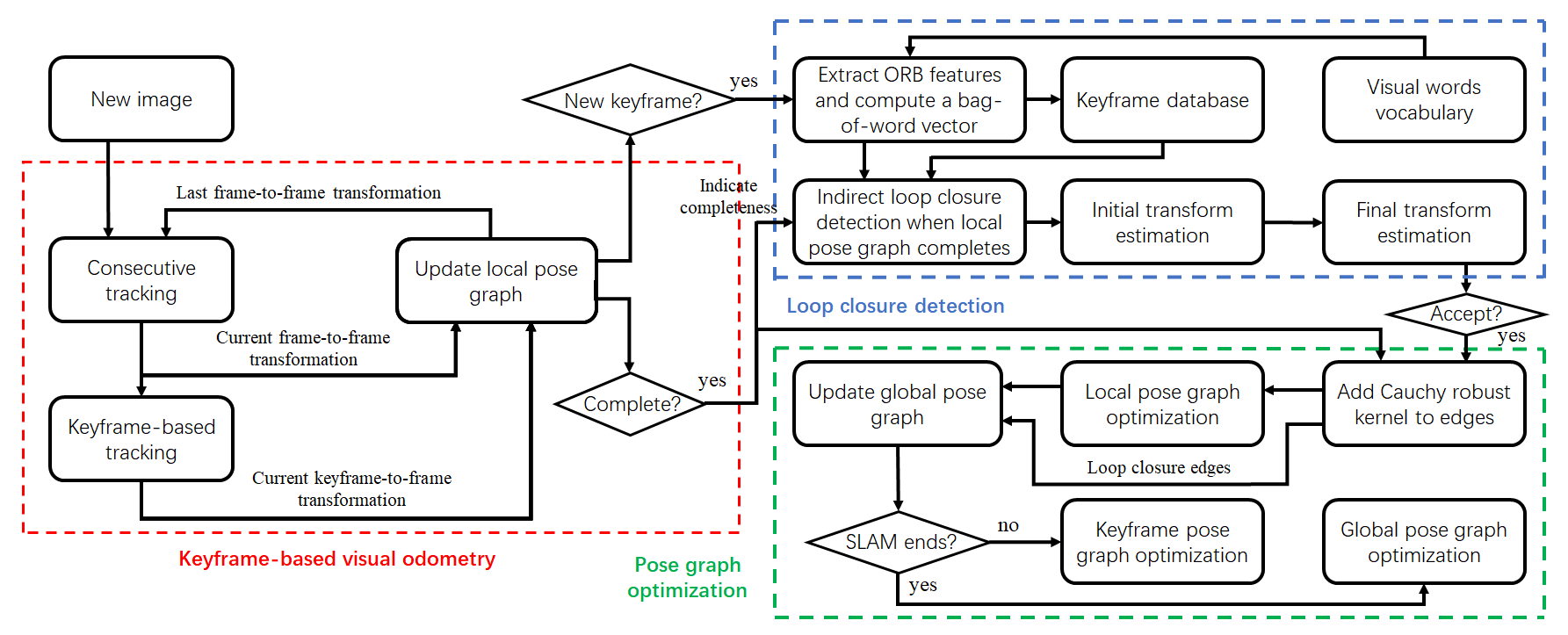}
   \caption{Pipeline for CVO SLAM. When a new scan is fed into the system, both frame-to-frame tracking and keyframe-based tracking will be performed. The resulting transformations will then be added to the local pose graph. A pose graph optimization is performed on the local pose graph before it is inserted into the global pose graph. When a new keyframe is decided, ORB features will be extracted from the frame, and a bag-of-words vector will be computed accordingly for future loop-closure detection. The loop-closure detection will be triggered whenever a local pose graph is added to the global pose graph. Then Cauchy robust kernel is added to all edges in current local pose graph and loop-closure edges. When the SLAM process finishes, a final optimization will be performed on the global pose graph.}
\label{fig:slam}
\end{figure*}

To correct accumulated tracking drift, we perform indirect loop-closure detection among keyframes using ORB features~\cite{orb}. We obtain an initial pose estimate between any loop-closure keyframes using matched point clouds that are generated by matched features and their corresponding depth values. Then, we use our continuous visual odometry initialized by the previously estimated pose to compute a more accurate loop-closure pose. A schematic pipeline for loop-closure detection is shown in Figure~\ref{fig:slam} (The part inside the blue dash line box).

\subsection{Loop-closure detection via ORB features}

When a new keyframe is selected, following the corresponding steps in ORB-SLAM2~\cite{murORB2} and bag-of-words definition in~\cite{GalvezTRO12}, we extract ORB features from its image and compute a bag-of-words vector for it. Once the current local pose graph in tracking is complete and inserted to the global pose graph, loop-closure detection between the current keyframe and other keyframes except the last keyframe is performed; because the transformation from the last keyframe to current keyframe is already computed in keyframe-based visual odometry.

To measure the similarity between the images of current keyframe and other keyframes, following the definition in~\cite{GalvezTRO12} and repeated for convenience here in~\eqref{eq:similarity_score}, we compute the $\ell_1$-score $s(v_i,v_j)$ between the bag-of-words vector of the current keyframe $v_i$ and that of each previous keyframe except the last keyframe as  
\begin{equation}\label{eq:similarity_score}
      s(v_i,v_j):=1-\frac{1}{2}\left |\frac{v_i}{|v_i|}-\frac{v_j}{|v_j|}\right| \quad j=1,\dots,i-2
\end{equation}

The range of these scores is related to query images and the distribution of words in bag-of-words vectors. Thus we use normalized scores $\eta(v_i,v_j)$~\cite{GalvezTRO12}. Since the indirect method is only used in loop-closure detection between keyframes, bag-of-words vectors are kept only for keyframes.
We expect $v_{i-1}$, the bag-of-words vector of the keyframe just before the current keyframe, to be the most similar to $v_i$ and give the best score $s(v_i,v_{i-1})$. Therefore, we compute $ \eta(v_i,v_j)=\frac{s(v_i,v_j)}{s(v_i,v_{i-1})}$ 
and set a fixed threshold to accept all previous keyframes that have greater $\eta$ than the threshold to be loop-closure candidates.

\subsection{Loop-closure transformation estimation}

For each candidate keyframe, we match its ORB features to those of the current keyframe using the feature matching method in ORB-SLAM2~\cite{murORB2} and perform the RANSAC algorithm to get inlier matches. Since we have access to RGB-D images, we can directly get the corresponding 3D points of the matched features. The rigid body transformation between these two sets of points with known correspondences can be computed using a singular value decomposition (SVD) as described in~\cite{SVD_estimation}, which serves as an initial pose, $T_j^i$, from the current keyframe $i$ to the previous keyframe $j$. This approach has been named as sparse-to-dense, first proposed in~\cite{dai2017bundlefusion}.

Since the number of features extracted in indirect methods is limited, the point clouds used to compute an initial pose are often sparse. To get a more accurate estimate of the pose, we use our continuous visual odometry, which is a direct method and has good alignment quality for close and similar images. We note that the similarity of the two images was established by the indirect detection method described earlier. The final loop-closure pose is equivalent to solving the following problem. 
\begin{equation}\label{eq:loop closure CVO}
    T_{\text{CVO}}=(\argmax_{h\in \mathrm{SE}(3)} \,  \langle f_{X_i}, h.f_{X_j}\rangle_{\mathcal{H}})^{-1},\ h_0 = T_i^j.
\end{equation}

Before we decide to add a loop-closure edge into the global pose graph, we conduct quality control. In keyframe-based odometry, we use the inner product of CVO functions as a proxy for the alignment quality. We also use it here and define a parameter $\alpha$ as follows.
\begin{equation}\label{eq:loop closure quality}
\begin{split}
    \alpha := \langle f_{X_i},&(T_{CVO})^{-1}.f_{X_j}\rangle_{\Hcal} \\
    &-\max(
    \begin{bmatrix}
    \langle f_{X_i},f_{X_j}\rangle_{\Hcal} \\
    \langle f_{X_i},T_i^j.f_{X_j}\rangle_{\Hcal} \\
    \langle f_{X_i},(T_{\text{rel}})^{-1}.f_{X_j}\rangle_{\Hcal} \\
    \end{bmatrix}
    )
\end{split}
\end{equation}
where $\langle f_{X_i},f_{X_j}\rangle_{\Hcal}$ corresponds to the case where two frames are identical, $T_{\text{rel}}=T_{\text{W}}^i T_j^{\text{W}}$, and $T_i^{\text{W}}$ and $T_j^{\text{W}}$ are the global pose estimates for two loop-closure keyframes. If $\alpha > 0$, it means that $T_{CVO}$ has better image alignment quality in the sense of CVO inner product than other transformations. Therefore, we accept the loop-closure for incorporation into the global pose graph. 

\section{Robust pose graph optimization}
\label{sec:posegraph}

After new loop-closures are added to the global pose graph, we optimize the pose graph to correct the accumulated error of the trajectory. Figure~\ref{fig:slam} shows this process (inside green dash line box). Since the size of the global pose graph grows fast, it would be inefficient if we frequently performed optimization on the entire graph. Instead, we extract keyframe nodes and edges between them for the optimization.

We first perform optimization on the current local pose graph with its keyframe node being set as fixed temporarily, updating the pose estimates of ordinary frames with respect to that of the keyframe node. This is equivalent to solving~\eqref{eq:local pose graph optimization}, where $\mathcal{P}$ is the set of all poses in the local pose graph, $e_{ij}$ is the error (residual) between pose $i$ and $j$ estimates and their relative pose measurement $T_{ij}$. 
\begin{equation}\label{eq:local pose graph optimization}
    \argmin_{\mathcal{P}} \, \sum_{(i,j)\in\mathcal{P}} e_{ij}^\transpose\Omega_{ij}e_{ij}.
\end{equation}
where $\Omega_{ij}$ is the information matrix of edge $ij$ in the pose graph.

Then, the fixed condition for the current node is removed, and we optimize all keyframe nodes and edges between them, which is called a keyframe pose graph optimization in Figure~\ref{fig:slam}. However, there is still a problem here. Although we try to accept high-quality loop-closures, the conditions are not sufficient, and outliers can exist, which can affect the optimization result. As such, before new loop-closures are used in the optimization, we add a Cauchy robust kernel, using g2o library~\cite{g2o}, where the Cauchy loss function is $\rho(x) = \delta^2 \log(\frac{x}{\delta^2}+1)$.
$\delta$ can be tuned to adjust where the function starts becoming sublinear. $\rho(x)$ weakens the influence of large error terms and makes optimization more robust to outliers.

\begin{table}[t]
\centering
\caption{ Parameters used in experiments. Best match feature threshold in the table is the ratio of descriptor distance between a feature and its best match feature and its second-best match feature. The purpose of our experiments is to understand how well CVO SLAM can perform. Thus $\eta_{thres}$ and the minimum match number threshold are set to relatively low values to accept sufficient loop-closure constraints.\label{tab:parameters}}
\footnotesize
\begin{tabular}{lcr}
\toprule
    Parameters & Symbol & Value \\
        \midrule
        Translation threshold & $t_{\text{thres}}$ & $0.15 \m$ \\
        Rotation angle threshold & $\theta_{\text{thres}}$ & $30\degree$ \\
        Inner product ratio threshold & $\gamma_{\text{thres}}$ & $0.7$ \\
        Normalized similarity score threshold & $\eta_{\text{thres}}$ & $0.3$ \\
        Best match feature threshold & & $0.7$ \\
        Minimum match number threshold &  & $5$ \\
        Robust kernel delta value & $\delta$ & $2$  \\
        Kernel sparsification threshold & & $8.315\mathrm{e}{-3}$ \\
        Spatial kernel initial length-scale & $\ell_{init}$ & $0.1$ \\
        Spatial kernel signal variance & $\sigma$ & $0.1$  \\
        Color kernel length-scale & $\ell_c$ & $0.1$ \\
        Color kernel signal variance & $\sigma_c$ & $1$  \\
\bottomrule
\end{tabular}
\label{tab:parameters}
\end{table}

\begin{table*}[t]
    \centering
    \caption{The RMSE of Absolute Trajectory Error (ATE) for \texttt{fr1} sequences. Since BAD SLAM failed to finish some \texttt{fr1} sequences in our experiment, Average* shows the average ATE RMSE only on sequences that BAD SLAM did not fail. Average all is the average ATE RMSE for all sequences shown in the table. Maximum shows the result of sequence with maximum ATE RMSE. There's no corresponding validation datasets for \texttt{fr1/teddy}. The results show that CVO SLAM has the lowest Average* values on training sequences and second lowest on validation sequences. \label{tab:fr1_SLAM_result}}
    \resizebox{\textwidth}{!}{
    \begin{tabular}{lcccc|cccc}
        \toprule
         & \multicolumn{4}{c}{Training} & \multicolumn{4}{c}{Validation} \\
        \midrule
         \multicolumn{1}{l}{Sequence} 
          & \multicolumn{1}{c}{CVO SLAM} & \multicolumn{1}{c}{DVO SLAM~\cite{kerl2013dense}} & \multicolumn{1}{c}{ORB-SLAM2~\cite{murORB2}} & \multicolumn{1}{c}{BAD SLAM~\cite{schps2019bad}} & \multicolumn{1}{c}{CVO SLAM} & \multicolumn{1}{c}{DVO SLAM~\cite{kerl2013dense}} & \multicolumn{1}{c}{ORB-SLAM2~\cite{murORB2}} & \multicolumn{1}{c}{BAD SLAM~\cite{schps2019bad}} \\
        \midrule
        fr1/desk & 0.0251 & 0.0222 & \bf 0.0159 & 1.1037 & 0.0315 &	0.0274 & \bf 0.0197 & 0.6919 \\
        fr1/desk2 & 0.0342 & 0.0290 &	\bf 0.0229 & 0.0335 & 0.0176 & 0.0175 & \bf 0.0107 & 0.0217 \\ 
        fr1/room & 0.1108 & 0.0796 & \bf 0.0493 & 0.1664 & 0.0719 & 0.3351 & \bf 0.0213 & failed \\
        fr1/360 &  \bf 0.0659 & 0.0975 &	0.2333 & 0.1617 & \bf 0.0636 & 0.0788 &	0.1218 & 0.1779  \\
        fr1/teddy &  0.0591 & \bf 0.0395 & 0.0538 & failed & n/a & n/a & n/a & n/a \\
        fr1/xyz & 0.0167 & 0.0131 & \bf 0.0096 & 0.0172 & 0.0110 & 0.0080 & \bf 0.0067 & failed \\
        fr1/rpy & 0.0225 & 0.0233 & \bf 0.0216	& 0.0237 & 0.0542 & \bf 0.0256 & 0.0304 & 0.2003 \\
        fr1/plant & 0.0184 & 0.0299 &	\bf 0.0135 & 0.0587 & 0.0520 &	0.2372 & \bf 0.0208 & 0.4066\\
        \midrule
        Average* & \bf 0.0419 & 0.0421 & 0.0523 & 0.2236 & 0.0485 & 0.0773 & \bf 0.0407 & 0.2997 \\
        Average all & 0.0441 & \bf 0.0418 & 0.0525 & n/a & 0.0431 & 0.1042 & \bf 0.0331 & n/a \\
        Maximum & 0.1108 & \bf 0.0975 & 0.2333 & failed & \bf 0.0719 & 0.3351 & 0.1218 & failed \\
        \bottomrule
    \end{tabular}
    }
    \label{tab:fr1_results}
\end{table*}

\begin{table*}[t]
    \centering
    \caption{The RMSE of Absolute Trajectory Error (ATE) for the structure v.s texture sequence. Since ORB-SLAM2 failed to complete some sequences, Average* shows the average ATE RMSE only on sequences that ORB-SLAM2 did not fail. Average all is the average results for all sequences. Maximum is the largest ATE RMSE among all sequences. The \checkmark means the sequence has structure/texture and $\times$ means the sequence does not have structure/texture. The results show that CVO SLAM has the lowest Average all value, second lowest Average* value and the lowest Maximum ATE RMSE on both training and validation sequences.\label{tab:fr3_SLAM_result}}
    \resizebox{\textwidth}{!}{
    \begin{tabular}{lllcccc|cccc}
        \toprule
         & & & \multicolumn{4}{c}{Training} & \multicolumn{4}{c}{Validation} \\
        \midrule
           \multicolumn{3}{c}{Sequence} & \multicolumn{1}{c}{CVO SLAM} & \multicolumn{1}{c}{DVO SLAM~\cite{kerl2013dense}} & \multicolumn{1}{c}{ORB-SLAM2~\cite{murORB2}} & \multicolumn{1}{c}{BAD SLAM~\cite{schps2019bad}} & \multicolumn{1}{c}{CVO SLAM} & \multicolumn{1}{c}{DVO SLAM~\cite{kerl2013dense}} & \multicolumn{1}{c}{ORB-SLAM2~\cite{murORB2}} & \multicolumn{1}{c}{BAD SLAM~\cite{schps2019bad}} \\
         \multicolumn{3}{l}{structure-texture-dist.} \\
        \midrule
        $\times$   & \checkmark & near & \bf 0.0198 & 0.0491 & 0.0236 & 0.0247 & 0.0197 & 0.0665 & \bf 0.0155 & 0.0392 \\
        $\times$   & \checkmark & far & \bf 0.0291 & 0.0847 & 0.0341 & 0.0799 & \bf 0.0293 & 0.0625 & 0.0294 &	0.1547  \\
        \checkmark & $\times$   & near & 0.0272 & 0.0726 & failed & \bf 0.0101 & 0.0275 & \bf 0.0176 & failed & 0.0633 \\
        \checkmark & $\times$   & far & 0.0491 & \bf 0.0432 & failed & 0.0943 & 0.0177 & 0.0213 & failed & \bf 0.0170\\
        \checkmark & \checkmark & near & 0.0333 & 0.0431 & 0.0129 & \bf 0.0137 & 0.0476 & 0.0334 & \bf 0.0116 & 0.0150 \\
        \checkmark & \checkmark & far & 0.0252 & 0.0199 & \bf 0.0109 & 0.0240 & 0.0322 & 0.0233 & \bf 0.0134 & 0.0279 \\
        $\times$   & $\times$   & near & \bf 0.2083 & 1.6169 & failed & 1.6644 & \bf 0.2302 & 1.6613 & failed & 1.6605 \\
        $\times$   & $\times$   & far & \bf 0.1538 & 0.8061 & failed & 0.8006 & \bf 0.1371 & 0.9275 &	failed & 0.9600 \\
        \midrule
        
		\multicolumn{3}{c}{Average*} & 0.0269 & 0.0492 & \bf 0.0204 & 0.0356 & 0.0322 & 0.0464 & \bf 0.0175 & 0.0592   \\
		\multicolumn{3}{c}{Average all} & \bf 0.0682 & 0.3420 & n/a & 0.3390 & \bf 0.0677 & 0.3517	& n/a &	0.3672 \\
	    \multicolumn{3}{c}{Maximum} & \bf 0.2083 & 1.6169 & failed & 1.6644 & \bf 0.2302 & 1.6613 & failed & 1.6605 \\
        \bottomrule
    \end{tabular}
     }
    \label{tab:fr3_results}
\end{table*}

\begin{figure*}[t]
\begin{center}
\includegraphics[width=2\columnwidth]{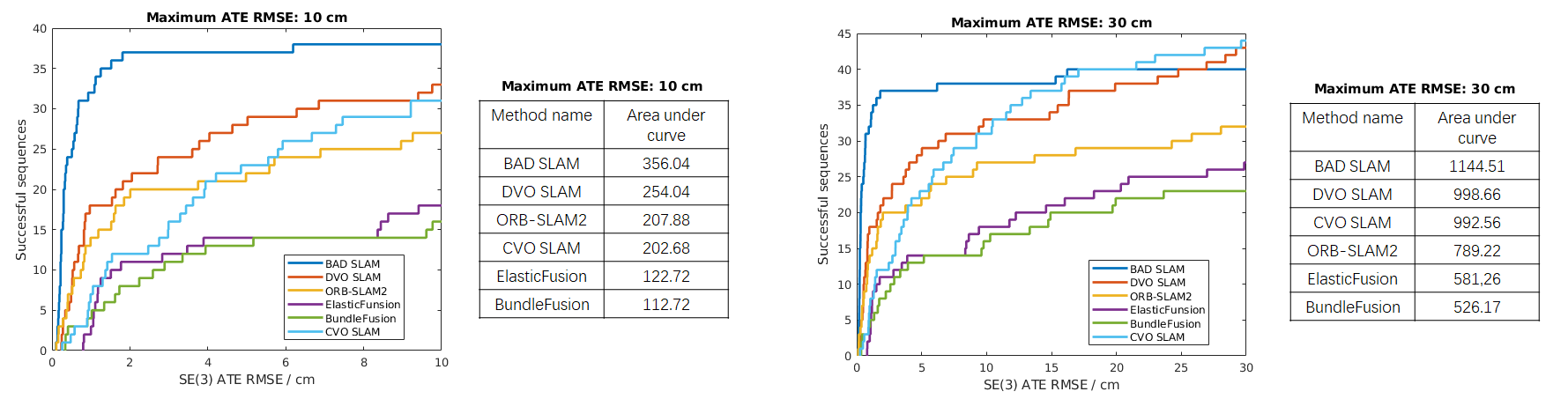}
\end{center}
\caption{Cumulative error visualization plot for CVO SLAM and SLAM benchmark results on training sequences of the ETH3D RGB-D dataset. ETH3D dataset and SLAM benchmark results for several RGB-D SLAM methods that we use here are proposed by Schöps \textit{et al.}~\cite{schps2019bad}. We plot the cumulative error of CVO SLAM as light blue curves and compute the area under curves for evaluation. When the maximum ATE RMSE error is set to 10 cm, the overall performance of CVO SLAM is as good as ORB-SLAM2 in terms of area under the curve. When maximum ATE RMSE error is set to 30 cm, the overall performance of CVO SLAM is as good as DVO SLAM.}
\label{fig:ETH3D_training_results}
\vspace{-2mm}
\end{figure*}

\section{Experimental results}
\label{sec:exp}

To evaluate the performance of CVO SLAM, we conduct experiments using TUM RGB-D dataset~\cite{TUM_dataset} and ETH3D RGB-D dataset~\cite{schps2019bad}. We also run DVO SLAM~\cite{kerl2013dense}, ORB-SLAM2~\cite{murORB2}, and BAD SLAM~\cite{schps2019bad} for comparison. Since CVO~\cite{MGhaffari-RSS-19} is a semi-dense method, we followed the point selection approach proposed in Direct Sparse Odometry (DSO)~\cite{engel2018direct}. The point selection process selects around 3000 pixels per image and creates a semi-dense point cloud. When point clouds are being generated, normalized gradients of the gray-scale image are also computed and, along with the RGB values, are used as the appearances of the point cloud. These appearances are then utilized as the labels in~\eqref{eq:newscalar}. 

All experiments are performed on a Lenovo Y700 laptop with Intel i7-6700HQ CPU (4 cores with 2.60 GHz each) and 16GB RAM. The parameters, shown in Table~\ref{tab:parameters}, are only tuned on \texttt{fr1} training sequences of the TUM RGB-D dataset and used for all experiments, i.e., the parameters remained the same across all experiments. Since BAD SLAM is brittle on the TUM dataset, we use the best out of three successful trials (ignoring failed cases) as its final outcome.

\subsection{TUM RGB-D SLAM dataset: Freiburg1}

In this section, we presents results on the \texttt{fr1} sequences of TUM RGB-D dataset~\cite{TUM_dataset}. The \texttt{fr1} sequences are recorded indoor with sufficient texture and structure in the environment. Because of some missing frames in the original video of \texttt{fr1/floor}~\cite{Whelan2016elastic}, we exclude this sequence. Table~\ref{tab:fr1_SLAM_result} shows the Root-Mean-Squared Error (RMSE) of the Absolute Trajectory Error (ATE) of the four SLAM methods. ORB-SLAM2 has the best performance across the sequences. This is reasonable since ORB-SLAM2 extracts ORBs from the image and maintains a map as part of its back-end graph. ORBs are expected to work well in environments that have rich textures. However, compared with DVO SLAM, which is also a direct method, CVO SLAM shows similar performance on the training sequences and has an overall lower average RMSE error on the validation sequences.

\subsection{Structure vs. texture: Freiburg3}

Table~\ref{tab:fr3_SLAM_result} shows the experimental results using \texttt{fr3} sequences in TUM RGB-D dataset~\cite{TUM_dataset} which are recorded in structure/nostructure and texture/notexture environments. The result shows that CVO SLAM has a better overall performance \emph{without tracking failure}. ORB-SLAM2 failed in cases without texture, which is expected. As for sequences that ORB-SLAM2 succeeded to complete, CVO SLAM also achieved a smaller average ATE RMSE than that of DVO SLAM and BAD SLAM. For the two most challenging sequences, namely no structure and no texture near and far, CVO SLAM significantly outperforms all the compared baselines. This experiment shows the robustness of CVO SLAM by verifying the ability to perform under no structure or no texture environment without failing on both training and validation sets.

\subsection{ETH3D RGB-D dataset}

To evaluate the performance of CVO SLAM under different situations, we also ran it on all 61 training sequences of the ETH3D RGB-D dataset proposed by Schöps \textit{et al.}~\cite{schps2019bad}. Since the parameters we used were only tuned in \texttt{fr1} training sequences of the TUM dataset~\cite{TUM_dataset}, the ETH3D training sequences serve as test sequences here. Figure~\ref{fig:ETH3D_training_results} shows the cumulative error plot for the results of CVO SLAM and other RGB-D SLAM systems. This plot is introduced in the benchmark results of ETH3D datasets. Although the error curve of CVO SLAM does not grow fast at the beginning, indicating that the number of sequences with very small ATE RMSE error is low, it keeps growing steadily afterward. Hence it shows that the number of successful sequences with an error smaller than a given threshold keep growing steadily as the threshold increases. The area under the CVO SLAM error curve was roughly the same as that of ORB-SLAM2 and DVO SLAM when maximum ATE RMSE error is set to 10 cm and 30 cm, respectively, showing that CVO SLAM has equivalent performance.   

\subsection{Discussions and limitations}

From the experiments and comparison with the state-of-the-art SLAM systems, the developed CVO SLAM shows promising and, overall, good performance in different indoor domains and under challenging camera motion regimes. The current results show that CVO SLAM is robust to severe cases where no structure or texture present in the scene. Besides, the same set of parameters works for different datasets, indicating it generalizes well across different domains.

Nevertheless, there are certain aspects of the current work that can be improved. First, ORB-SLAM2 and BAD-SLAM use bundle adjustment and maintain a map that can be used for localization. Since setting correct covariances for each edge in the pose graph is challenging, using the map in the graph for relocalization gives better performance. The current CVO SLAM framework only maintains a pose graph. In the future, we shall explore how to exploit a sparse and semi-dense map in the graph optimization directly. Another important factor to consider is that CVO is a locally optimal algorithm. As such, similar to other compared baselines in this work, the performance depends on the initial guess. We used the estimate from any registration to initialize the next registration, which is a common way of providing a warm start. We expect initialization using an inertial measurement unit can bring noticeable improvements to the system.

\section{Conclusion and future work}
\label{sec:conclusion}
We developed a robust keyframe-based RGB-D visual odometry that performs well in scarcely textured and structured environments and has better generalizability across different training and validation sequences. The proposed approach models RGB-D images using a nonparametric joint geometric and appearance representation in a Reproducing Kernel Hilbert Space (RKHS). We showed the alignment of two RGB-D images, as well as keyframe selections, that could be done using the inner product structure of the RKHS. We extended the developed keyframe-based odometry to a SLAM system using indirect ORB loop-closures and showed comparable performance with the state-of-the-art using publicly available datasets.

Given the promising results in this work, we consider the following topics as interesting future work. The addition of an inertial measurement unit can improve the initialization during tracking. Maintaining a sparse or semi-dense map in the back-end graph analogous to the bundle adjustment problem will make relocalization in the map possible and most likely improve the performance. We conjecture the latter using our experience in working with the ORB-SLAM2 system. Finally, our current implementation is not real-time; a real-time implementation of the developed keyframe-based continuous visual odometry on GPUs can bring robustness to visual front-end and SLAM systems.

{\small
\bibliographystyle{ieee_fullname}
\bibliography{strings-full,ieee-full,refs}
}
\end{document}